\documentclass{article}

\usepackage[nonatbib, final]{neurips_2021}

\usepackage[utf8]{inputenc} % allow utf-8 input
\usepackage[T1]{fontenc}    % use 8-bit T1 fonts
\usepackage{hyperref}       % hyperlinks
\usepackage{url}            % simple URL typesetting
\usepackage{booktabs}       % professional-quality tables
\usepackage{amsfonts}       % blackboard math symbols
\usepackage{nicefrac}       % compact symbols for 1/2, etc.
\usepackage{microtype}      % microtypography
\usepackage{xcolor}         % colors
\usepackage{amsmath}
\usepackage{float}
\usepackage{graphicx}
\usepackage{caption}
\usepackage{subfigure}
\usepackage{cite}

\newcommand\blfootnote[1]{%
  \begingroup
  \renewcommand\thefootnote{}\footnote{#1}%
  \addtocounter{footnote}{-1}%
  \endgroup
}

\title{NoFADE: Analyzing Diminishing Returns on CO2 Investment}

\author{
  Andre Fu \thanks{University of Toronto} \thanks{Corresponding Author: \texttt{andre.fu@utoronto.ca}}\\
  \And
  Justin Tran \footnotemark[1] \\
  \And
  Andy Xie \footnotemark[1]\\
  \And 
  Jonathan Spraggett\footnotemark[1] \\
  \And
  Elisa Ding \footnotemark[1]\\
  \And
  Chang-Won Lee\footnotemark[1]\\
  \And
  Kanav Singla\footnotemark[1]\\
  \And
  Mahdi S. Hosseini\thanks{University of New Brunswick, \texttt{mahdi.hosseini@unb.ca}}\\
  \And
  Konstantinos N. Plataniotis\footnotemark[1]\\
}

\begin{document}
% get rid of subsections

\maketitle

\begin{abstract}
  % The abstract paragraph should be indented \nicefrac{1}{2}~inch (3~picas) on both the left- and right-hand margins. Use 10~point type, with a vertical spacing (leading) of 11~points.  The word \textbf{Abstract} must be centered, bold, and in point size 12. Two line spaces precede the abstract. The abstract must be limited to one paragraph.
  Climate change continues to be a pressing issue that currently affects society at-large. It is important that we as a society, including the Computer Vision (CV) community take steps to limit our impact on the environment. In this paper, we (a) analyze the effect of diminishing returns on CV methods, and (b) propose a \textit{``NoFADE''}: a novel entropy-based metric to quantify model--dataset--complexity relationships. We show that some CV tasks are reaching saturation, while others are almost fully saturated. In this light, NoFADE allows the CV community to compare models and datasets on a similar basis, establishing an agnostic platform.
\end{abstract}

\section{Introduction}
We are currently experiencing a surge in the development and implementation of neural architectures in the computer vision (CV) community. This is accompanied by exponential increases in both model complexity and accuracy, which are consistently correlated with performance dependent on computer power \cite{strubell2019energy}. The compute required for training AI models follows an exponential trend with a 3.4 month doubling period \cite{amodei_2021}, as state of the art (SOTA) models (eg. NAS, AlphaZero) utilize over $\times{0.3}$M the compute of previous benchmarks (eg. AlexNet). The community is heavily focused on surpassing SOTA results. However, this will require further increases in compute power and downstream implications cannot be overlooked, particularly CO2 output of deep learning workloads.

%The compute required for training the biggest AI models has been doubling approximately every 3.4 months \cite{amodei_2021}, breaking Moore's Law 2-year hardware density doubling by a huge factor. The current focus has been on surpassing state-of-the-art (SOTA) results  where utilizing increasing amounts of compute to achieve this goal means a severe overlook of the downstream implications, in particular CO2 output from computationally heavy deep learning workloads.

The Intergovernmental Panel on Climate Change (IPCC) provides transparent reports on the effects of climate change to policy makers and the scientific community. In the 2019 IPCC report, the IPCC highlighted that the $1.5^{\circ}$C above pre-industrial levels is required to mitigate extreme climate related events. In 2016, there was 52 GtCO2 with an expected 52-58 GtCO2 by 2030, with an IPCC recommended 25-30 GtCO2 \cite{ipcc2021} in the same time frame. Recently, the sixth IPCC report was released where they stressed that ``global warming of $1.5^{\circ}$C  and $2.0^{\circ}$C will be exceeded in the 21st century unless deep reductions in CO2 and other greenhouse gases occur in the coming decades'' \cite{ipcc2021}. The climate crisis is a pressing issue that faces society at-large, and as such everyone, even those within the deep-learning community have their role to play to curb the climate crisis. It is crucial our community gains a better understanding of how our deep learning models affect the climate crisis, our responsibilities within the crisis, and improvements in our methodologies to include awareness of the crisis.

%begin to understand where we can improve our methods to include awareness of the climate crisis.
There has been no previous analysis on the topic within the deep learning and CV communities; therefore we propose and investigate the following questions:
%Within the deep learning community, specifically the CV  community no such analysis has been conducted, therefore this paper aims to do so; in an effort to address the following questions:
\begin{enumerate}
    \item How much CO2 is emitted compared to evaluation metrics for any model? Is there a saturation occurring within popular fields?
    \item How can we gain a better understanding of the relationships between models and datasets on the basis of CO2 investment?
\end{enumerate}
In answering these questions, we find that (1) there exists a diminishing return on increasing computation and equivalently CO2 emission and (2) we develop a metric to characterize the relationship between a model and the dataset's difficulty of learning while normalizing for complexity.
 
\section{Diminishing Returns of Increasing Computation}
%As deep learning explodes in popularity, there is a rush to compete against the state-of-the-art (SOTA) methods. 
As deep learning grows in popularity, there is corresponding competition to improve upon SOTA methods. In doing so, the community often overlooks the implicit costs of chasing SOTA such as the increasing computational burdens \cite{strubell2019energy, fu2021reconsidering} which in turn have severe environmental consequences. These models necessitate growing computational footprints during training but also during their lifetime \cite{fu2021reconsidering} as deployed systems. Here, we analyze the trade-offs between increasing computational cost in Watt-hours and environmental cost in CO2 emissions. 

%Here, we analyze the trade-offs between increasing SOTA and the increasing computational and equivalently the environmental cost in CO2 emitted. 

% \subsection{Methodology} % increase this methodology
Inspired by \cite{fu2021reconsidering, strubell2019energy} methodologies, we begin by generating a corpus of computer vision models within the three most common tasks (a) classification, (b) segmentation and (c) detection. We surveyed 13 classification papers \cite{brock2021high, tan2021efficientnetv2, chen2019muffnet, ridnik2003tresnet, gao2019lip, ma2020funnel, Sandler_2018_CVPR, wang2020eca, tan2019mnasnet, ma2018shufflenet, li2019selective,dosovitskiy2020image,touvron2021going}, 22 segmentation papers  \cite{li2019expectation, park2018c3, paszke2016enet, hu2020efficient, olimov2021fu, ronneberger2019u, porzi2019seamless, gamal2018shuffleseg, yu2018bisenet, park2019semantic, mehta2018espnet, wang2020dual, li2019dfanet, huang2019ccnet, vallurupalli2018efficient, chen2020multi, he2019dynamic, zhuang2019shelfnet, kaul2021focusnet++, li2020humans, li2020micronet, he2019knowledge}  and 10 detection papers \cite{qin2019thundernet, carion2020end, ren2015faster, redmon2018yolov3, li2019light, lin2017focal, chen2019detnas, bochkovskiy2020yolov4, zhang2018single, liu2018receptive}. For every model, we extracted the Top-1 Test-accuracy, mAP or mIOU, FLOPS, GPU hours and GPU type. 

Using GPU type, the Watt-to-FLOPS was calculated using the model's FLOPS as \emph{f} as $\omega$ = Watt/\emph{f}. Then, the GPU Watt-to-FLOPS are denoted as {$\omega_g$} and the CPU's Watt-to-FLOPS as {$\omega_c$} \cite{fu2021reconsidering}. We obtain a model’s power draw, \emph{$P_m$} over training by multiplying the model’s flops, \emph{f} by the sum of the Watt-to-FLOPS ratios. This quantity would be multiplied by the total GPU hours to train \cite{fu2021reconsidering}, seen below:
% \emph{$P_m$}[Wh] = $\emph{f} \times ({$\omega_g$} + {$\omega_c$})$ $ \times $ GPU Hours
\begin{equation}
    P_m \text{[Wh]}= f \times ( \omega_g + \omega_c) \times \text{GPU hours}.
    \label{eq:flop_pow}
\end{equation}

The final CO2 emissions are calculated by multiplying the power draw from a model by the EPA’s Wh to CO2 measurement $0.707 \times 10^{-3}$ metric tonnes/kWh, CO2 = \emph{$P_m$} · $0.707 \times 10^{-3}$.
% \subsection{Results \& Discussion}
\begin{figure}[H]
% \begin{center}
    \centering
    % \hspace*{\fill}%
    \subfigure[ImageNet]{\includegraphics[width=0.35\textwidth]{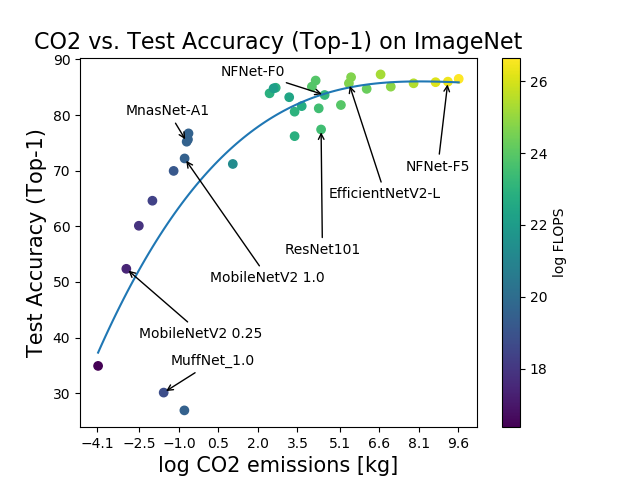}}\hspace{-1cm}
    \subfigure[COCO]{\includegraphics[width=0.35\textwidth]{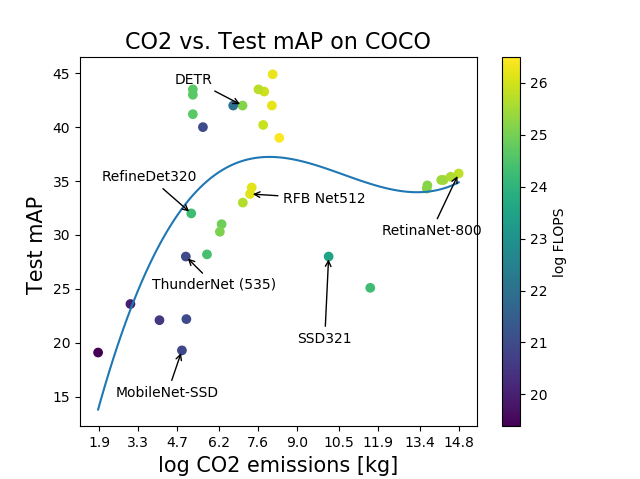}} \hspace{-1cm}
    \subfigure[PASCAL]{\includegraphics[width=0.35\textwidth]{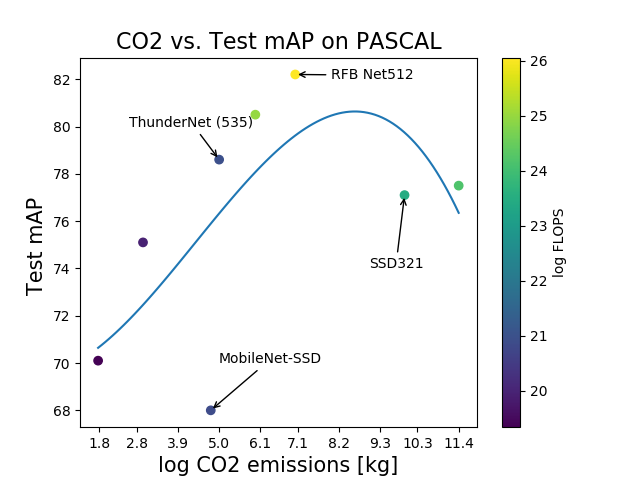}} 
    \subfigure[CityScapes]{\includegraphics[width=0.35\textwidth]{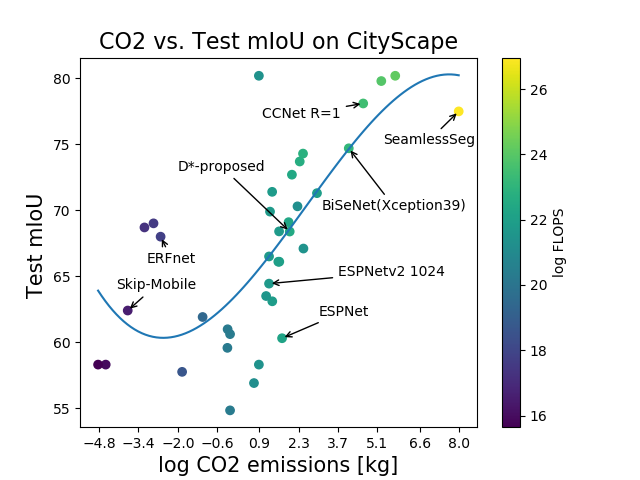}} \hspace{-1cm}
    \subfigure[PASCAL VOC]{\includegraphics[width=0.35\textwidth]{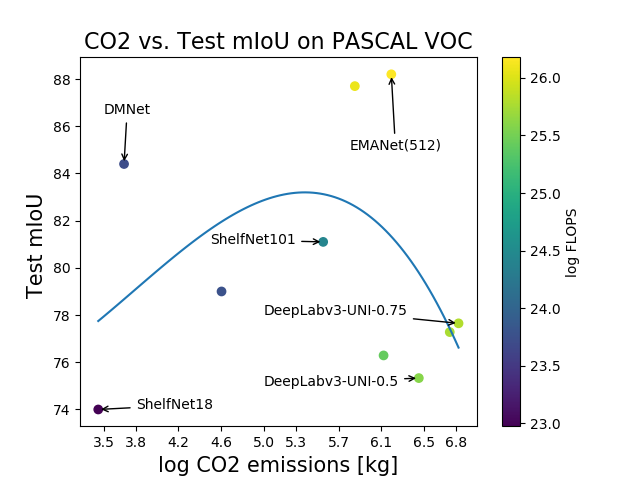}} 
    \caption{(a) Classification task ImageNet \cite{krizhevsky2012imagenet}, (b, c) Sematic Segmentation tasks MS COCO \cite{lin2014microsoft} and PASCAL \cite{everingham2010pascal}, (d,e) Detection tasks CityScape \cite{cordts2016cityscapes} and PASCAL VOC \cite{everingham2010pascal}}
    \label{fig:co2_vs_dataset}
\end{figure}

% remove colour bar, only have 1 colour bar

%becomes clear why ImageNet has such a clear saturation, as the field is reaching an optima without significant improvements in recent years.
Within classification, the ImageNet dataset \cite{krizhevsky2012imagenet} has the most distinct saturation as seen in \autoref{fig:co2_vs_dataset}(a). This implies a clear diminishing return on `investment' of CO2 emissions for gained accuracy. The classification task is the most mature, combined with long-term interest prior to deep-learning methods. To that end, it is justified that ImageNet reaches saturation as the field reaches optimal performance with minimal improvements in recent years. In fields such as Segmentation and Detection that are still developing, the saturation curve is observable but less well defined, as in \autoref{fig:co2_vs_dataset}(b) and (d). Notice in all the subfigures the models with more FLOPS generally emit significantly more CO2, yet these high-FLOPS models have the same or lower performance in comparison to low-FLOPS models.
%most of these model do not outperform or sometimes even underperform compared with models that have less emissions. 
On MS COCO, PASCAL, and PASCAL VOC, we see that it is possible to trade-off computational complexity to achieve SOTA performance by reducing the FLOPS constraint. For example, implementing ThunderNet(535) in lieu of SSD321 would achieve better test performance with less FLOPS, thus reducing carbon emissions without sacrificing results.
%The most evident example of this would be SSD321 to ThunderNet(535). ThunderNet(535) achieves better test performance with less FLOPS, thus leading to less emissions.
 
\section{NoFADE Development}
% FOCUS ON THE MODEL & DATASET pair (MODEL_I, DATASET_J)
In the pursuit of increasing SOTA at a diminishing computational cost, we find that there currently is no concrete analysis of the dataset learning space in relationship to models. To identify if a model is learning well on a dataset, the difficulty of learning must be determined, and weighed with the relationship to test-accuracy. We begin by creating a collection of the unique model $i$ and dataset $j$ pairs ($\text{Model}_i, \text{Dataset}_j)$. For any two model-dataset pairs, it becomes difficult to compare their relative performances while taking into account difficulty of learning and computational complexity of a model. In this work, we attempt to characterize a novel metric, \textit{NoFADE}: Normalized FLOPS for Accuracy-Dataset Entropy which allows the CV community to compare models in an effective way. 

% \subsection{Methodology}\label{sec:entropy-meth}
We use Shannon entropy as the underlying measure of difficulty of learning \cite{rahane2020measures}. We begin by converting colour images to greyscale and using $
    H(\text{Image}) = - \sum_{i = 1}^{K} p_i \cdot \log p_i
    \label{eq:entropy} $ to determine the entropy. 
The Image being analyzed has a frequency of pixel intensity $p_i$, with $K$ unique intensities.

\textbf{Segmentation and Detection} Segmentation (Cityscapes, PASCAL) and Detection (MS COCO, PASCAL) datasets are comprised of a set of images with their reference true labels. As entropy represents the randomness within an image, we can assess the total randomness of the dataset. Therefore we apply Shannon Entropy to each image in the dataset to generate a normal distribution. As shown in \cite{rahane2020measures} measuring a dataset's entropy distribution allows us to gain a better understanding of image complexity and most importantly the difficulty of learning. 
% SHOW THE HISTOGRAMS OF MULTIPLE DATASETS (

\begin{figure}[H]
    % \begin{center}
    \centering

    \subfigure{\includegraphics[width=0.2\linewidth]{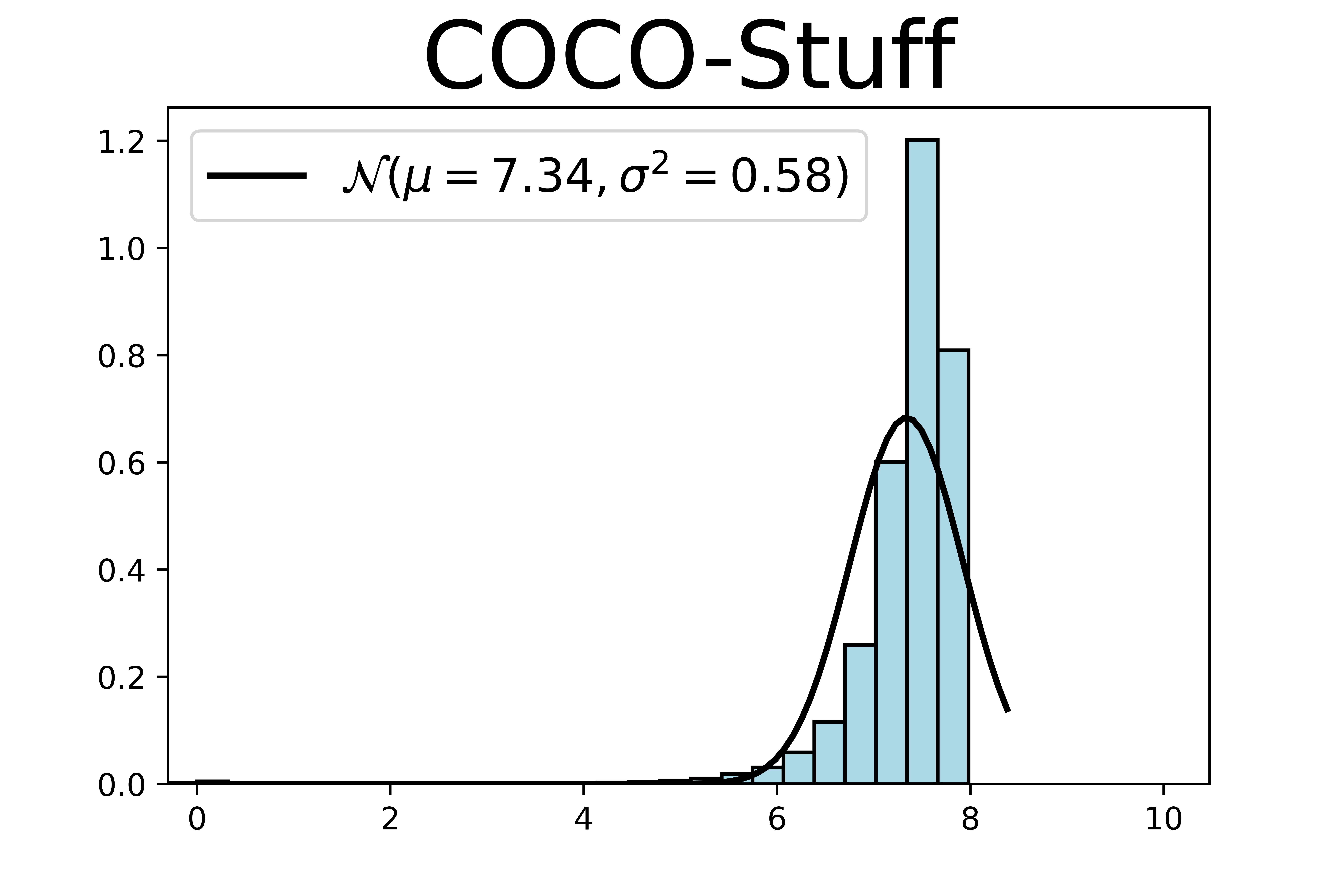}} \hspace{-0.4cm}
    \subfigure{\includegraphics[width=0.2\linewidth]{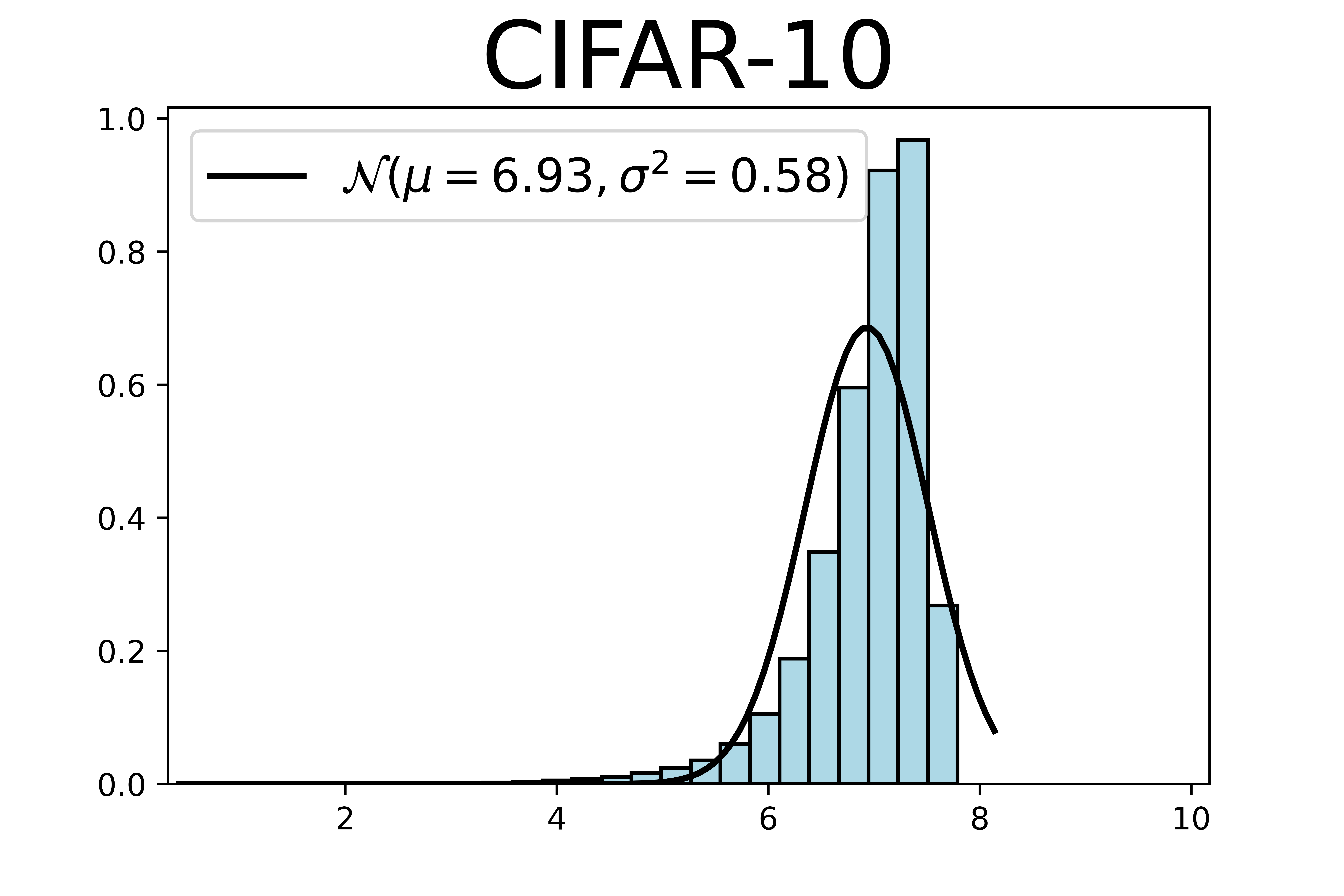}} \hspace{-0.4cm}
    \subfigure{\includegraphics[width=0.2\linewidth]{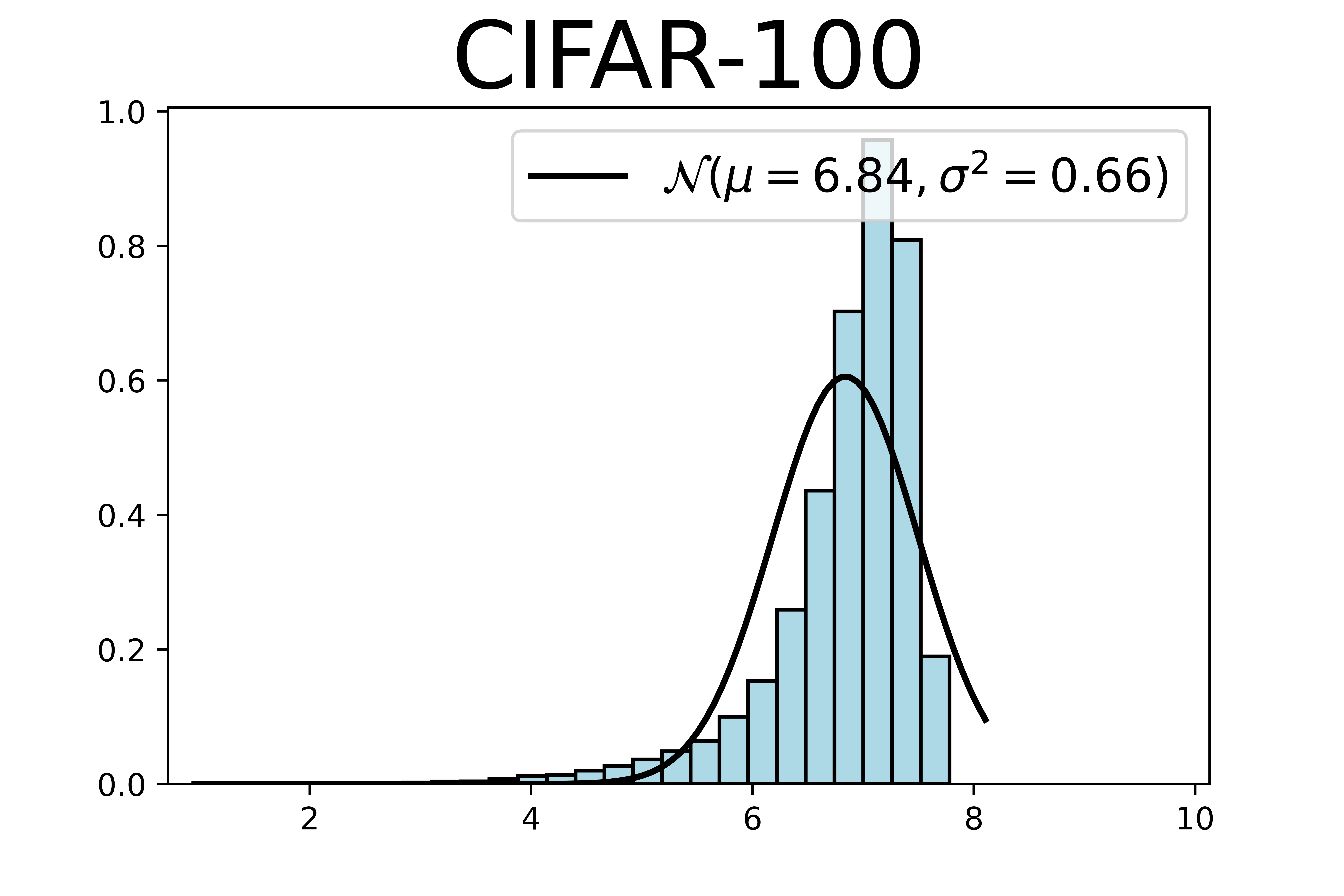}}\hspace{-0.3cm}
    \subfigure{\includegraphics[width=0.2\linewidth]{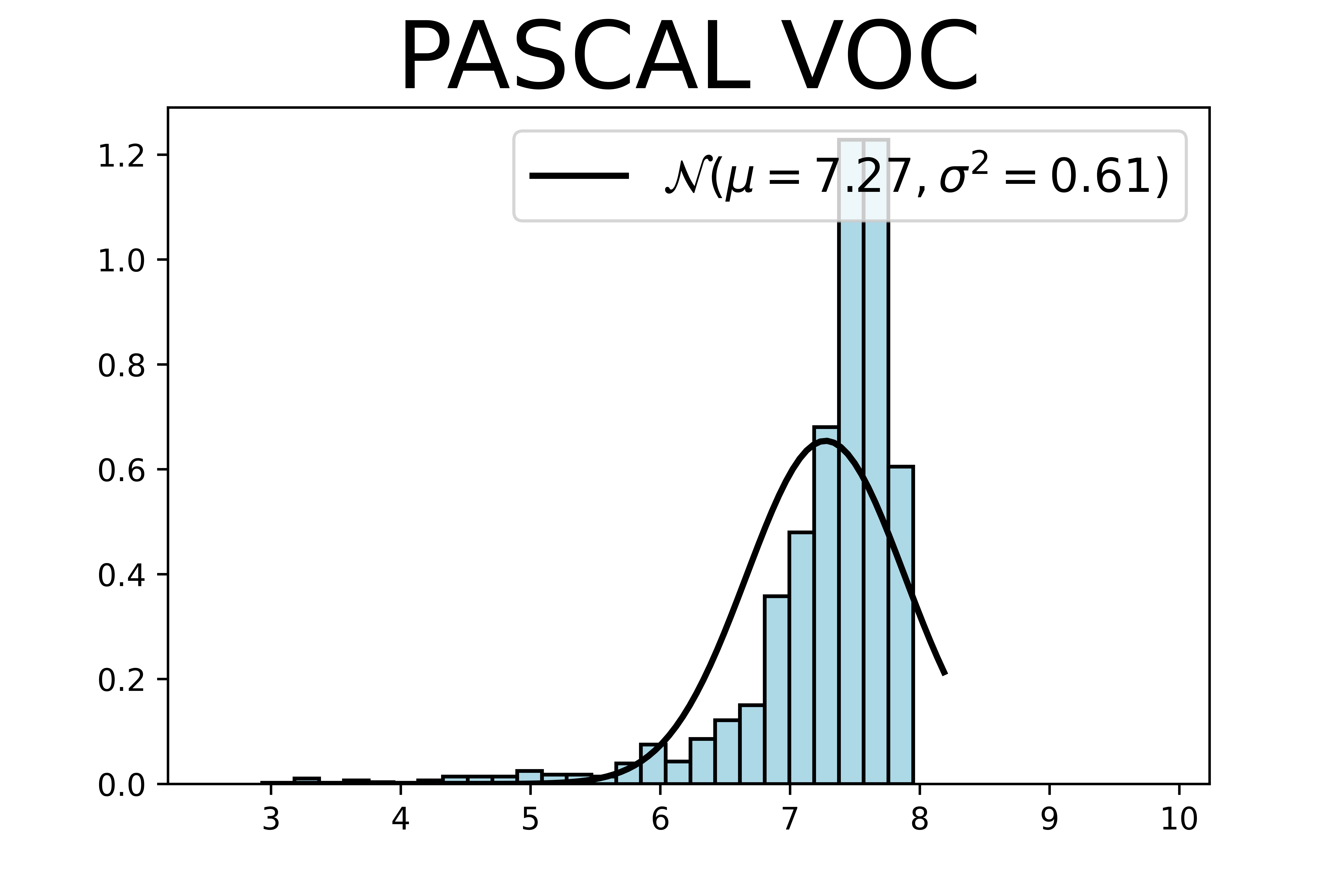}}\hspace{-0.3cm}
    \subfigure{\includegraphics[width=0.2\linewidth]{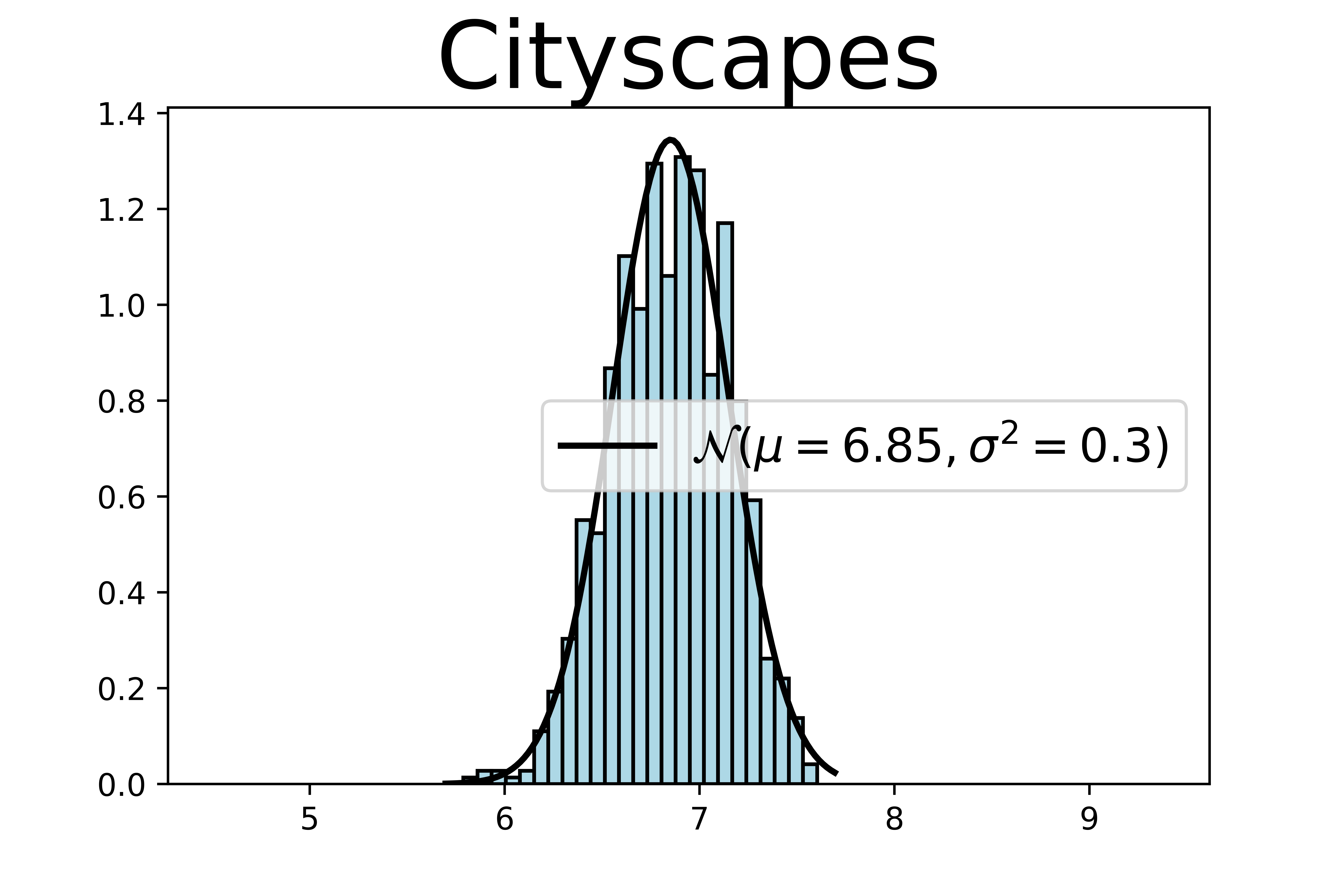}}
    % \end{center}
    \caption{Entropy Histograms for Cityscapes, Pascal VOC, CIFAR10, CIFAR100 and MS COCO}
    \label{fig:Histogram entropy}
\end{figure}

\textbf{Classification:} Classification difficulty is inherently related not only to image complexity, but also multi-distribution learning as the classes are themselves distributions. To resolve this, we compute the entropy using Shannon Entropy for every class within CIFAR10, CIFAR100 and ImageNet. Each class distribution now represents the complexity of that class, but to gain an informational relationship about the dataset as a whole, we can measure the distances between classes. To do this we employ the Jensen-Shannon distance \cite{fuglede2004jensen}, a statistical measure that assesses the distance between two distributions in a symmetric and finite manner. Based on the Kullback-Leibler divergence, Jensen-Shannon distance allows us to gain understanding of mutual information between the two distributions. 
\begin{equation}
    JSD(P \lVert Q) = \sqrt{\frac{1}{2} D_{KL}(P \lVert M) + \frac{1}{2} D_{KL}(Q \lVert M)}.
    \label{eq:jsd}
\end{equation}

Here, $M = \frac{1}{2}(P + Q)$ and $D_{KL}$ refers to the KL-divergence. Using \autoref{eq:jsd} we determine the distance between each class distributions, which represents the difficulty of learning between the two classes. Therefore, the sum of the Jensen-Shannon distances for the unique pairs of classes, provides a measure of the complexity of the dataset as a whole, with higher values representing more difficult learning. In order to normalize the data to a similar scale as the entropy calculations, we take the $\log$ of the summation.

NoFADE attempts to characterize a relationship between models and datasets while normalizing for computational complexity. To do this, we propose multiplying the test-accuracy gained by the entropy of the dataset. As a dataset's difficulty of learning is correlated to it's entropy or JS-distance, we are effectively increasing the amount of ``metric points'' a model is gaining in proportion to the difficulty of learning on that dataset. For example a $1\%$ increase on ImageNet is ``better'' in the community than a $1\%$ increase on CIFAR10. To that end, we also recognize that the informal rule of ``deeper performs better'' may skew these results. To mitigate this risk, we normalize the ``metric points'' by the $\log \text{FLOPS}$, as most of the FLOPS are tightly grouped the $\log$ allows us to exploit sensitivity. Doing so lets the metric understand that higher FLOP models have a constraint and thus deserve a normalization to allow for comparisons. 
\vspace{-0.2cm}
\begin{equation}
    \text{NoFADE} = \frac{\text{test-accuracy} \times \text{Entropy or JS-dist}}{\log \text{FLOPS}}
\end{equation}
\vspace{-1cm}
\begin{figure}[H]
% \begin{center}
    \centering
    \subfigure[Classification]{\includegraphics[width=0.34\linewidth]{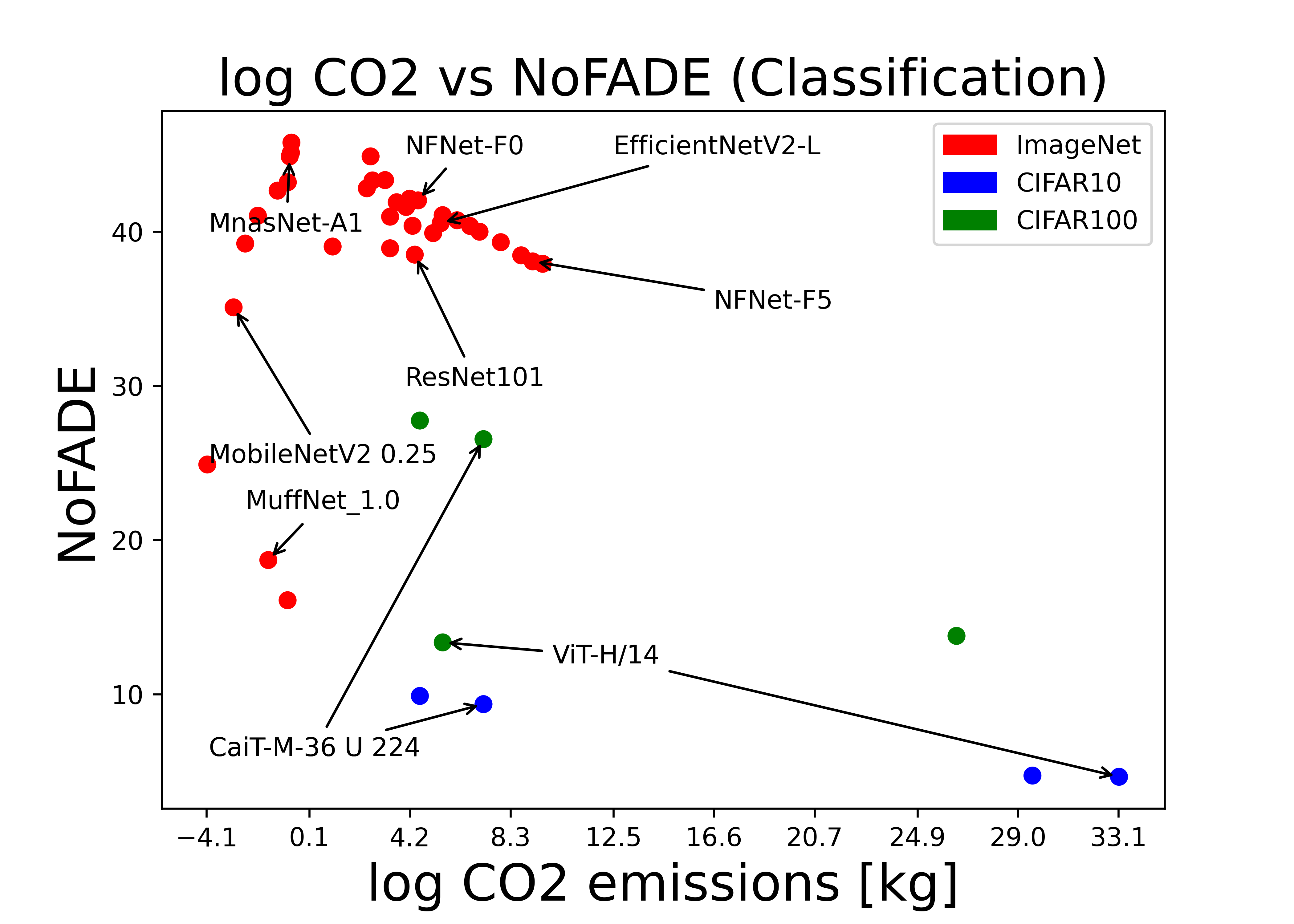}} \hspace{-0.5cm}
    \subfigure[Segmentation]{\includegraphics[width=0.34\linewidth]{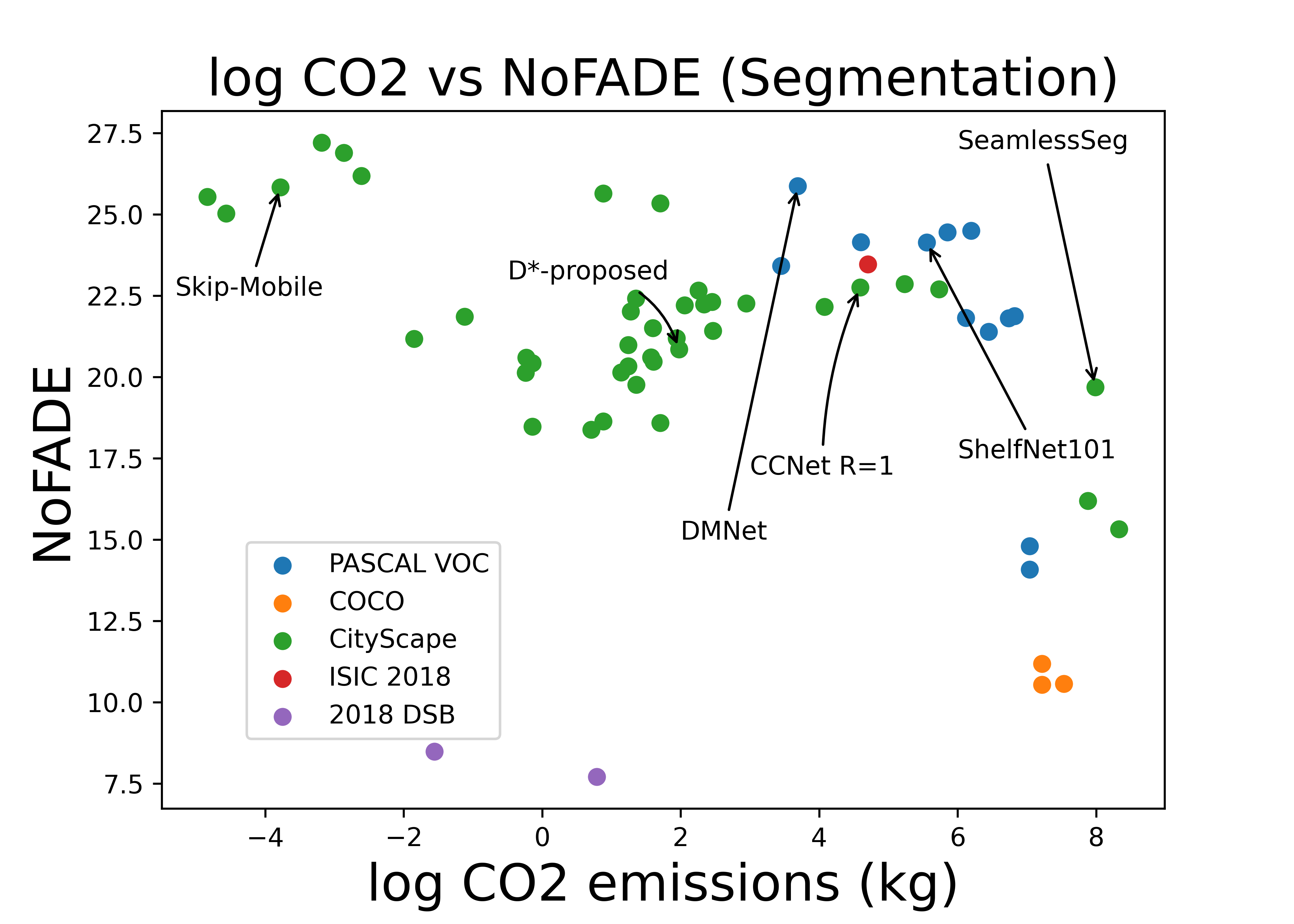}} \hspace{-0.6cm}
    \subfigure[Detection]{\includegraphics[width=0.34\linewidth]{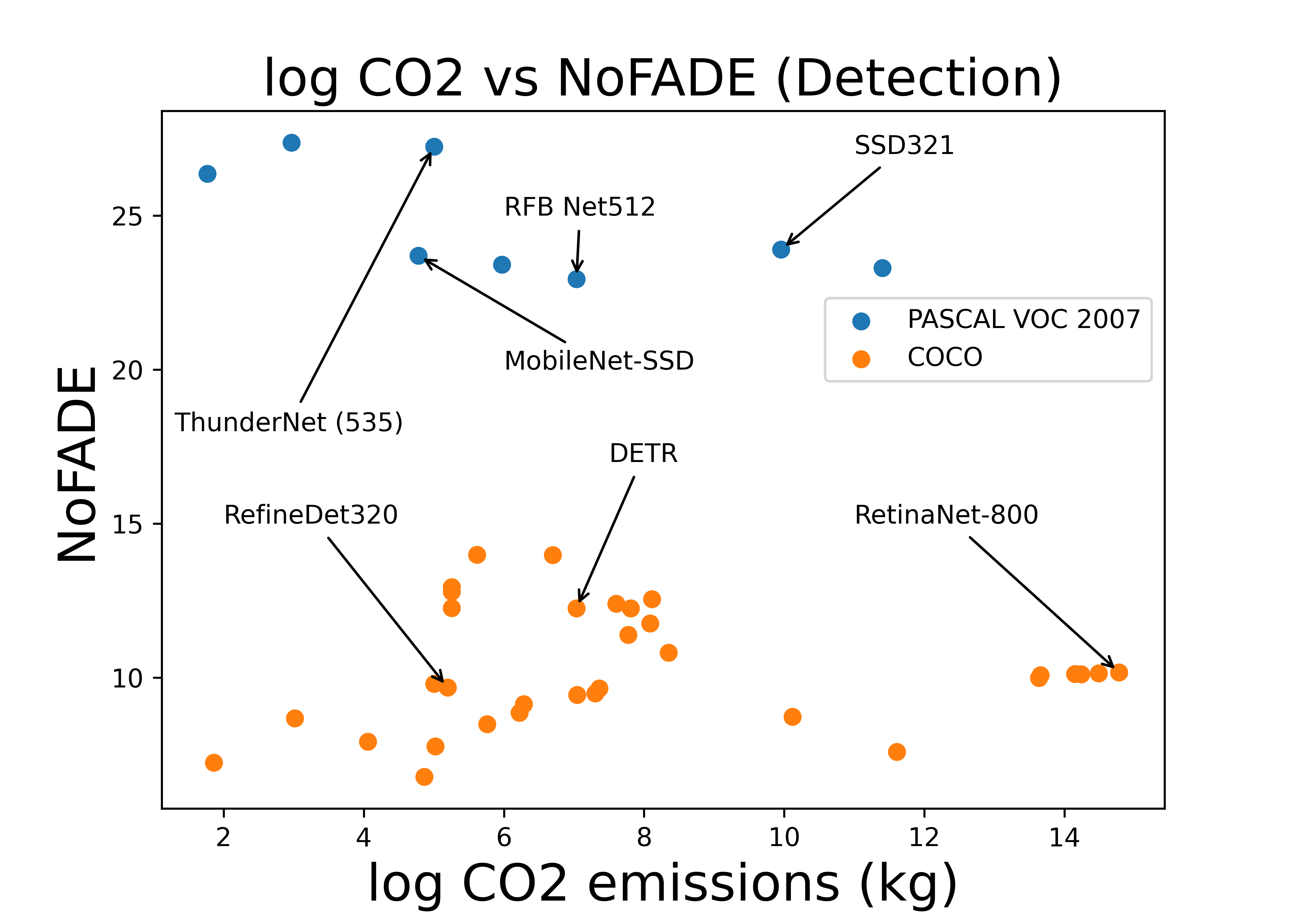}}
    % \end{center}
    \caption{$\log$ CO2 vs. NoFADE, (a) Classification with ImageNet, CIFAR10, CIFAR100 (b) Segmentation with COCO, PASCAL, Cityscape, ISIC 2018 and DSB 2018, (c) Detection with MS COCO and PASCAL VOC}
    \label{fig:nofade}
\end{figure}

In \autoref{fig:nofade} we can compare methods in a models--dataset--complexity--agnostic manner, building the foundation for comparisons. For example, because the ViT transformer models have such large FLOPS they do perform extremely well but at a significant complexity cost which is reflected in the lower NoFADE score, their training times are also reflected with their extremely large CO2 footprints. Therefore, implying the transformer models don't perform well on these key dimensions. Similarly, some of the best models like MNasNet perform extremely well in all three categories, having low CO2 while maintaining a high NoFADE score.

\section{Conclusion}
In this paper, we demonstrate how our choices as the CV community have an impact on global climate change. Specifically, the selection of various architecture and dataset pairs were explored and how dataset complexity affects a model's efficiency and performance, resulting in increased CO2 emissions. We propose a new metric to quantify a dataset's complexity and validated as a reasonable method to track CO2 emissions arising from model training. We hope that this proposal will help the AI community be more environmentally conscious and to be as efficient as possible when training their models and utilize the novel tool to measure model--dataset--CO2 relationship to gain a better understanding of their place in the field.  In addition, this measurement aid could facilitate model and dataset selection for performance and efficiency trade-off. Overall, with this new metric in hand, we anticipate researchers to start trying to reduce their environmental impact and hopefully find new ways to do so as well.\blfootnote{ We'd like to thank Michal Fishkin for her worthwhile contributions without whom this work would not have taken place.}

%%%%%%%%%%%%%%%%%%%%%%%%%%%%%%%%%%%%%%%%%%%%%%%%%%%%%%%%%%%%

{\small
\bibliographystyle{ieeetr}
\bibliography{egbib}
}

\end{document}